\documentclass{article} 
\usepackage{nips12submit_e,times}

\usepackage{amssymb}
\usepackage{graphicx}

\usepackage{subfigure} 

\usepackage[ruled]{algorithm2e}

\usepackage{hyperref}

\usepackage{amsmath}
\usepackage{amsfonts}
\usepackage{enumerate}

\newcommand{\bmat}{\begin{pmatrix}}
\newcommand{\emat}{\end{pmatrix}}

\newcommand{\Order}{\mathcal{O}}

\newcommand{\Normal}{\mathcal{N}}
\newcommand{\Expectation}{\mathbb{E}}

\newcommand{\popsize}{n}

\newcommand{\params}{\boldsymbol{\theta}}
\newcommand{\grad}{\nabla_{\params}}

\newcommand{\gradj}{\nabla_{\theta_{i}}}

\newcommand{\hatg}{\overline{g_i}}
\newcommand{\hatv}{\overline{v_i}}
\newcommand{\hath}{\overline{h_i}}

\newcommand{\citep}{\cite}

\newcommand{\adagrad}{{\sc AdaGrad}\ }

\usepackage{color}
\definecolor{dkgreen}{rgb}{0.1,0.4,0}
\definecolor{orange}{rgb}{0.8,0.4,0}
\definecolor{nyupurple}{rgb}{0.5,0.0,0.9}
\definecolor{violet}{rgb}{0.9,0.,0.5}

\title{Adaptive learning rates and parallelization for stochastic, sparse, non-smooth gradients}

\author{
Tom Schaul \\
Courant Institute of Mathematical Sciences\\
New York University, 715 Broadway,\\ 10003, New York\\
\texttt{schaul@cims.nyu.edu} 
\And Yann LeCun \\
Courant Institute of Mathematical Sciences\\
New York University, 715 Broadway, \\10003, New York\\
\texttt{yann@cims.nyu.edu} 
}

\nipsfinalcopy

\begin{document}

\maketitle





\begin{abstract}
Recent work has established an empirically successful framework
for \emph{adapting} learning rates for stochastic gradient descent (SGD).
This effectively removes all needs for tuning, while automatically
reducing learning rates over time on stationary problems, and
permitting learning rates to grow appropriately in non-stationary tasks.
Here, we extend the idea in three directions, addressing proper
minibatch parallelization, including reweighted updates for sparse or orthogonal gradients,
improving robustness on non-smooth loss functions, in the process
replacing the diagonal Hessian estimation procedure that may not always be available 
by a robust finite-difference approximation.
The final algorithm integrates all these components, has linear complexity and is hyper-parameter free.
\end{abstract}

\section{Introduction}

Many machine learning problems can be framed as minimizing a loss function
over a large (maybe infinite) number of samples. 
In representation learning, those loss functions are generally built on top of multiple layers of non-linearities, 
precluding any direct or closed-form optimization, but admitting (sample) gradients
to guide iterative optimization of the loss.

Stochastic gradient descent (SGD) is among the most broadly applicable and widely-used 
algorithms for such learning tasks, because of its simplicity, robustness and scalability to arbitrarily large datasets.
Doing many small but noisy updates instead of fewer large ones (as in batch methods) gives
both a speed-up, and makes the learning process less likely to get stuck in sensitive local optima.
In addition, SGD is eminently well-suited for learning in non-stationary environments, e.g., when 
that data stream is generated by a changing environment; but non-stationary adaptivity is useful 
even on stationary problems, as the initial search phase 
(before a local optimum is located) of the learning process can 
be likened to a non-stationary environment.

Given the increasingly wide adoption of machine learning tools, there is an undoubted benefit to
making learning algorithms, and SGD in particular, easy to use and hyper-parameter free.
In recent work, we made SGD hyper-parameter free by introducing 
optimal adaptive learning rates that are based on gradient variance estimates \citep{Schaul2012}.
While broadly successful, the approach was limited to smooth loss functions, and to minibatch sizes of one.
In this paper, we therefore complement that work, by addressing and resolving the issues of
\begin{itemize}
\item \emph{minibatches} and parallelization,
\item \emph{sparse} gradients, and
\item \emph{non-smooth} loss functions
\end{itemize}
all while retaining the optimal adaptive learning rates.
All of these issues are of practical importance: minibatch parallelization
has strong diminishing returns, but in combination with sparse gradients and adaptive learning rates, we show
how that effect is drastically mitigated. 
The importance of robustly dealing with non-smooth loss functions is also a very practical concern: 
a growing number of learning architectures employ non-smooth nonlinearities, like absolute value normalization
or rectified-linear units.
Our final algorithm addresses all of these, while remaining simple to implement and of linear complexity.

\section{Background}
\label{sec:vsgd}

There are a number of adaptive settings for SGD learning rates, or equivalently, diagonal preconditioning schemes, to be found in the literature, e.g., \cite{Jacobs1988,Almeida1999,George2006,NicolasLeRoux,DuchiHS11,amari2000adaptive}. The aim of those is generally to increase performance on stochastic optimization tasks, a concern complementary to our focus of producing an algorithm that works robustly without any hyper-parameter tuning. Often those adaptive schemes produce monotonically decreasing rates, however, which makes them no longer applicable to non-stationary tasks.

The remainder of this paper build upon the adaptive learning rate scheme of \citep{Schaul2012}, which is not monotonically decreasing,
so we recapitulate its main results here.
Using an idealized quadratic and separable loss function,
it is possible to derive an optimal learning rate schedule
which preserves the convergence guarantees of SGD.  
When the problem is approximately separable, 
the analysis is simplified as all quantities are one-dimensional.
The analysis also holds as a local approximation in the non-quadratic but smooth case.

In the idealized case, and for any dimension $i$, the optimal learning rate can be derived analytically, and
takes the following form
\begin{eqnarray}
\eta_i^* & =& \frac{1}{h_i} \cdot \frac{(\theta_i -\theta_i^*)^2} { (\theta_i -\theta_i^*)^2 + \sigma_i^2}
=
\frac{1}{h_i} \cdot \frac{\left(\Expectation [\nabla_{\theta_i}]\right)^2}
{\Expectation [\nabla_{\theta_i}^2]}
\label{eq:opt-lr}
\end{eqnarray}
where $(\theta_i-\theta^*_i)$ is the distance to the optimal parameter value, and $\sigma_i^2$ and $h_i$ are the local 
sample variance and curvature, respectively.

We use an exponential moving average with time-constant $\tau$ 
(the approximate number of samples considered from recent memory)
for online estimates of the quantities in equation~\ref{eq:opt-lr}:
\begin{eqnarray*}
\hatg  &\leftarrow& (1-\tau_i^{-1}) \cdot \hatg + \tau_i^{-1} \cdot \nabla_{\theta_i}\\
\hatv  &\leftarrow& (1-\tau_i^{-1}) \cdot \hatv + \tau_i^{-1} \cdot (\nabla_{\theta_i})^2\\
\hath  &\leftarrow& (1-\tau_i^{-1}) \cdot \hath + \tau_i^{-1} \cdot h_i^{(bbprop)}
\end{eqnarray*}
where the diagonal Hessian entries $ h_i^{(bbprop)}$ are computed using the `bbprop' procedure \citep{lecun-98b}, and the time-constant (memory) is adapted according to how large a step was taken:
\begin{eqnarray*}
\tau_i (t+1) &=& \left(1-\frac{\hatg(t)^2}{\hatv(t)}\right) \cdot  \tau_i(t) \;+\; 1 
\label{eq:memory-update}
\end{eqnarray*}

The final algorithm is called vSGD, and used the learning rates from
equation~\ref{eq:opt-lr} to update the parameters (element-wise):
\begin{eqnarray*}
\params  &\leftarrow& \params - \eta^* \cdot \grad
\end{eqnarray*}

\section{Parallelization with minibatches}
\label{sec:mb}

\begin{figure*}
	\centering
		\includegraphics[width=0.99\textwidth]{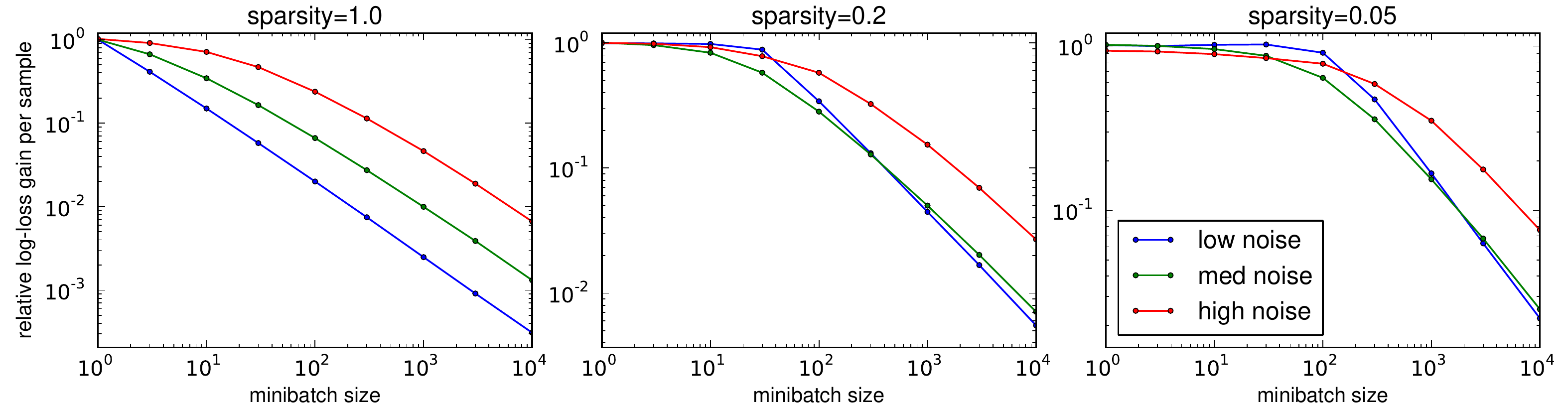}
\caption{Diminishing returns of minibatch parallelization.
Plotted is the relative log-loss gain (per number of sample gradients evaluated)
of a given minibatch size compared to the gain of the $\popsize=1$ case
 (in the noisy quadratic scenario from section \ref{sec:vsgd}, for different noise levels $\sigma$, and assuming optimal learning rates as in equation~\ref{eq:opt-lrsp});
each figure corresponds to a different sparsity level.
For example, the ratio is 0.02 for $\popsize=100$ (left plot, low noise): This means that it takes 50 times more samples to obtain the same gain in loss than with pure SGD.
Those are strongly diminishing returns, but they are less drastic if the noise level is high (only 5 times more samples in this example).
If the sample gradients are somewhat \emph{sparse}, however, and we use that fact to increase learning rates appropriately, then the diminishing returns kick in only for much larger minibatch sizes; see the left two figures.
}
	\label{fig:mbreturns}
\end{figure*}

Compared to the pure online SGD, computation time can be reduced
by ``minibatch''-parallelization: $\popsize$ sample-gradients are computed 
(simultaneously, e.g., on multiple cores) and
then a single update on the resulting averaged minibatch gradient is performed.
\begin{eqnarray}
\grad = \frac{1}{\popsize} \sum_{k=1}^{\popsize} \grad^{(k)}
\label{eq:mb}
\end{eqnarray}

While $\popsize$ can be seen as a hyperparameter of the algorithm~\cite{Byrd2012},
it is often constrained to a large extent by the computational 
hardware, memory requirements and communication bandwidth.
A derivation just like the one that led to equation~\ref{eq:opt-lr}
can be used to determine the optimal learning rates automatically,
for an arbitrary minibatch size $\popsize$.
The key difference is that the averaging in equation~\ref{eq:mb}
reduces the effective variance by a factor $\popsize$,
leading to:
\begin{eqnarray}
\eta_i^* & =& \frac{1}{h_i} \cdot \frac{(\theta_i -\theta_i^*)^2} { (\theta_i -\theta_i^*)^2 + \frac{1}{\popsize}\sigma^2}
=
\frac{1}{h_i} \cdot \frac{\left(\Expectation [\nabla_{\theta_i}]\right)^2}
{\frac{1}{\popsize}\Expectation [\nabla_{\theta_i}^2] + \frac{\popsize- 1}{\popsize}\left(\Expectation [\nabla_{\theta_i}]\right)^2}
\label{eq:opt-lrmb}
\end{eqnarray}
This expresses the intuition that using minibatches reduces the sample noise, in turn permitting
larger step sizes: if the noise (or sample diversity) is small, those gains are minimal, if it is large, they are substantial (see Figure~\ref{fig:mbreturns}, left).
%
Varying minibatch sizes tend to be impractical\footnote{If the implementation/computational architecture is flexible enough, the variance-term of the learning rate can also be used to adapt the minibatch size adaptively to its optimal trade-off.} 
to implement however, and so common practice is to simply fix a minibatch size, and then re-tune the learning rates (by a factor between 1 and $\popsize$).
With our adaptive minibatch-aware scheme (equation \ref{eq:opt-lrmb}) this is no longer necessary: in fact, we get an automatic transition from initially small effective minibatches (by means of the learning rates) to large minibatches toward the end, when the noise level is higher.

\section{Sparse gradients}
\begin{figure*}
	\centering
		\includegraphics[width=0.99\textwidth]{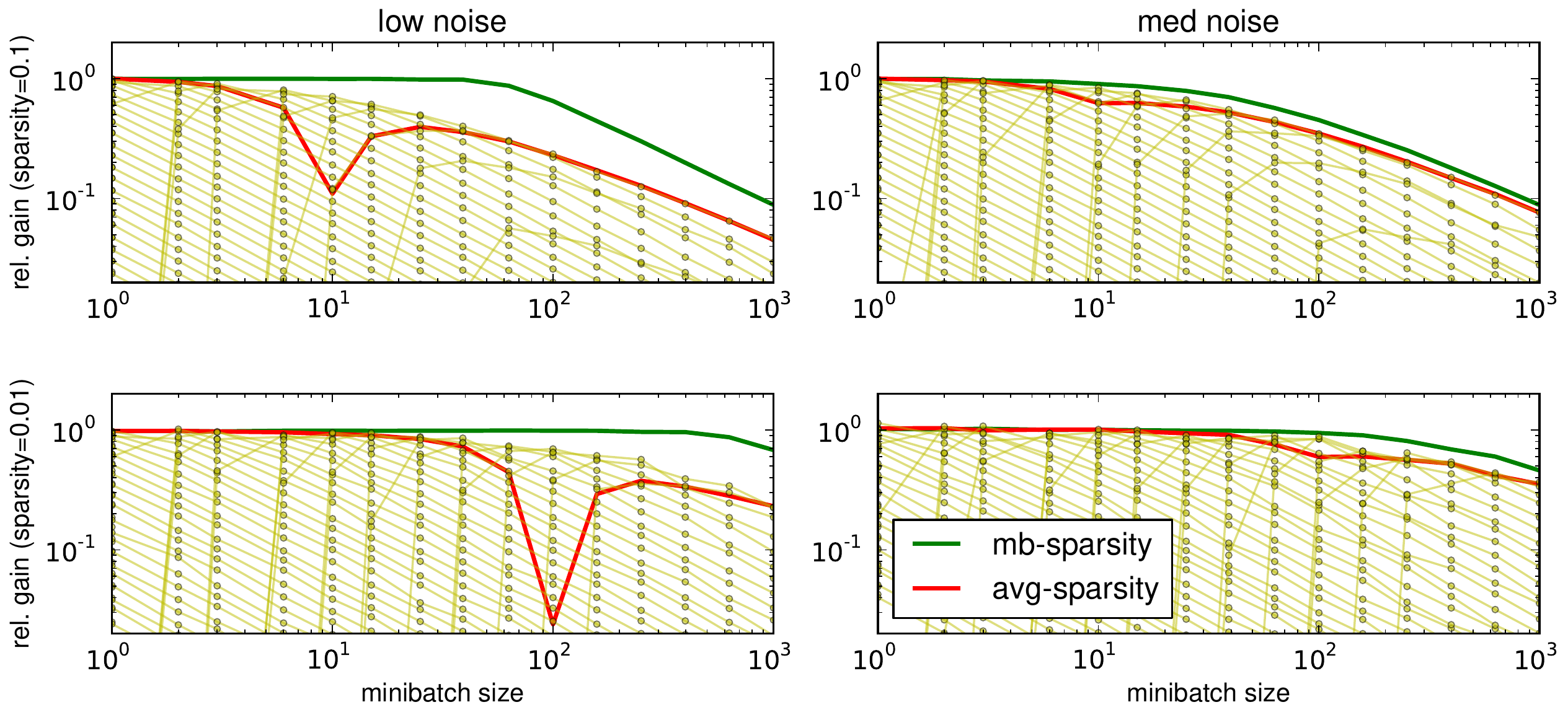}
\caption{Difference between global or instance-based computation of effective minibatch sizes
in the presence of sparse gradients.
Our proposed method computes the number of non-zero entries ($\popsize-z_i$) in the current mini-batch to set 
the learning rate (green). This involves some additional computation compared to just using the long-term average sparsity $p^{(nz)}_i$ (red), but obtains a substantially higher relative gain (see figure~\ref{fig:mbreturns}), especially in the regime where the sparsity level produces mini-batches with just one or a few non-zero entries (dent in the curve near $\popsize = 1/p^{(nz)}_i$).
If the noise level is low (left two figures), the effect is much more pronounced than if the noise is higher.
For comparison, the performance for 40 different fixed learning-rate SGD settings (between 0.01 and 100) are plotted as yellow dots.
}
	\label{fig:sparseoracles}
\end{figure*}

Many common learning architectures (e.g., those using rectified linear units, or sparsity penalties) lead to sample gradients that
are increasingly \emph{sparse}, that is, they are non-zero only in small fraction of the problem dimensions.
It is possible to exploit this to speed up learning, by averaging many sparse gradients in a minibatch, 
or by doing asynchronous updates~\citep{hogwild11}.

Here, we investigate how to set the learning rates in the presence of sparsity,
and our result is simply based on the observation
that doing an update using a set of sparse gradients is equivalent
to doing the same update, but with a smaller \emph{effective} minibatch size, while ignoring all the 
zero entries.

We can do this again on an element-by-element basis, where we define
$z_i$ to be the number of non-zero elements in dimension $i$,
within the current minibatch.
In each dimension, we rescale the minibatch gradient accordingly by a factor $\popsize/(\popsize-z_i)$,
and at the same time reduce the learning rate 
to reflect the smaller effective minibatch size.
Compounding those two effects gives the optimal learning rate for
sparse minibatches (we ignore the case $z_i=\popsize$, when there is no update):
\begin{eqnarray}
\eta_i^* & =& \frac{\popsize}{\popsize-z_i} \cdot \frac{1}{h_i} \cdot \frac{\left(\Expectation [\nabla_{\theta_i}]\right)^2}
{\frac{1}{\popsize-z_i}\Expectation [\nabla_{\theta_i}^2] + 
\frac{\popsize-z_i- 1}{\popsize-z_i}\left(\Expectation [\nabla_{\theta_i}]\right)^2}
\label{eq:opt-lrsp}
\end{eqnarray}

Figure~\ref{fig:mbreturns} shows how using minibatches with such adaptive learning 
rates reduces the impact of diminishing returns if the sample gradients are sparse.
In other words, with the right learning rates, higher sparsity can be 
directly translated into higher parallelizability.

An alternative to computing $z_i$ for each minibatch (and each dimension) anew
would be to just use the long-term average sparsity $p^{(nz)}_i = \Expectation\left[\popsize-z_i\right]$ instead. Figure~\ref{fig:sparseoracles} shows that this is suboptimal, especially if the noise level is small,
and in the regime where each minibatch is expected to contain just a few non-zero entries.
This figure also shows that equation~\ref{eq:opt-lrsp} produces a higher relative gain compared to the outer envelope of the performance of all fixed learning rates.

\subsection{Orthogonal gradients}

\begin{figure*}
	\centering
		\includegraphics[width=0.95\textwidth]{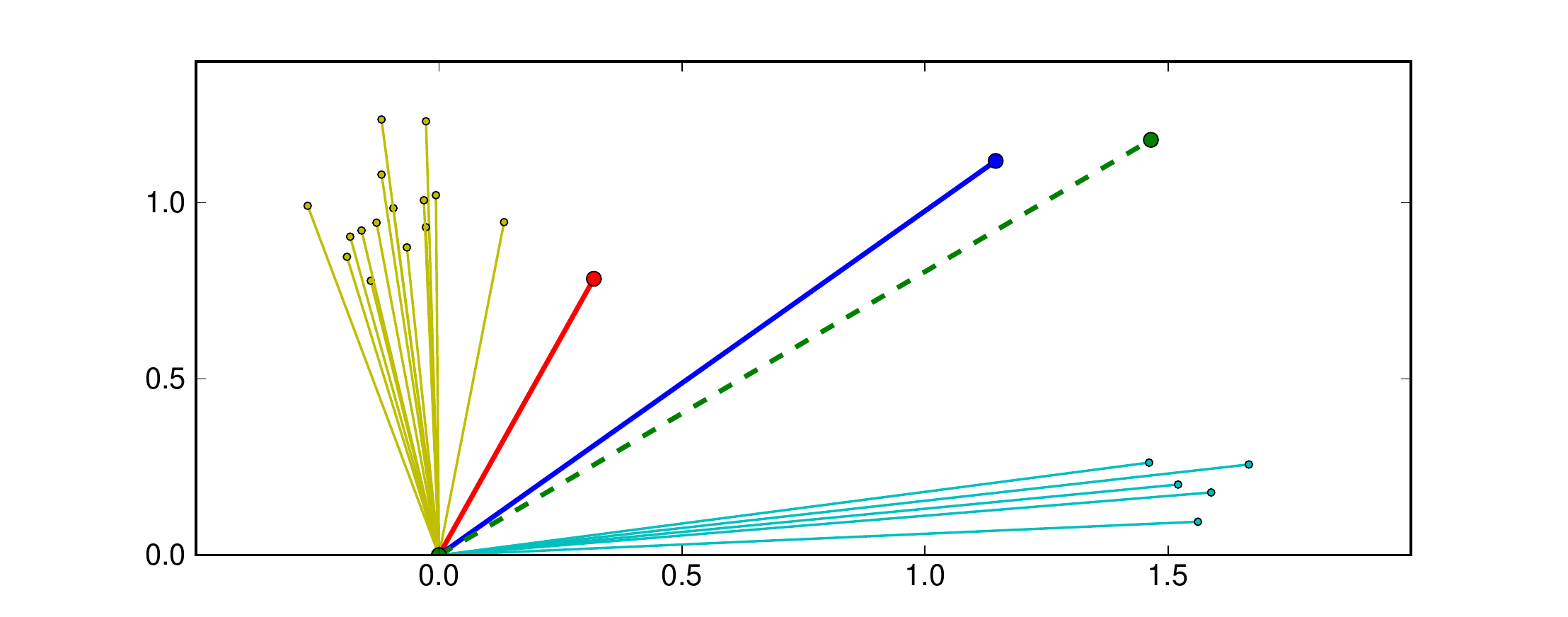}
\caption{Illustrating the effect of reweighting minibatch gradients.
Assume the samples are drawn from 2 different noisy clusters (yellow and light blue vectors),
but one of the clusters has a higher probability of occurrence.
The regular minibatch gradient is simply their arithmetic average (red), dominated by the more common cluster.
The reweighted minibatch gradient (blue) does a full step toward each of the clusters, 
closely resembling the gradient one would obtain by performing a hard clustering (difficult in practice) on the samples, in dotted green.
}
	\label{fig:orth-illust}
\end{figure*}

One reason for the boost in parallelizability if the gradients are sparse
comes from the fact that sparse gradients are mostly orthogonal,
allowing independent progress in each direction.
But sparse gradients are in fact a special case of
orthogonal gradients, for which we can obtain similar speedups
with a reweighting of the minibatch gradients:
\begin{eqnarray}
\grad &=& \sum_{i=1}^{\popsize} 
\frac{1}{\sum_{j=1}^{\popsize}  
\frac{|\grad^{(i)\top}  \grad^{(j)}|}
{\| \grad^{(i)} \| \cdot \| \grad^{(j)} \|} } 
\grad^{(i)} 
\label{eq:weightgrad}
\end{eqnarray}
In other words, each sample is weighted by one over the number of times (smoothed) that its gradient is interfering (non-orthogonal) with another sample's gradient.

In the limit, this scheme simplifies to the sparse-gradient cases discussed above: if all sample
gradients are aligned, they are averaged (reweighted by $1/\popsize$, corresponding to the dense case in equation~\ref{eq:mb}), and if all sample gradients
are orthogonal, they are summed (reweighted by 1, corresponding to the maximally sparse case $z_i=\popsize-1$ in equation~\ref{eq:opt-lrsp}). See Figure~\ref{fig:orth-illust} for an illustration.

In practice, this reweighting comes at a certain cost, increasing the computational expense of a single iteration from $\Order(nd)$ to $\Order(n^2d)$, where $d$ is the problem dimension. In other words, it is only likely to be viable if the forward-backward passes of the gradient computation are non-trivial, or if the minibatch size is small.

\section{Non-smooth losses}

\begin{figure*}
	\centering
		\includegraphics[width=0.99\textwidth]{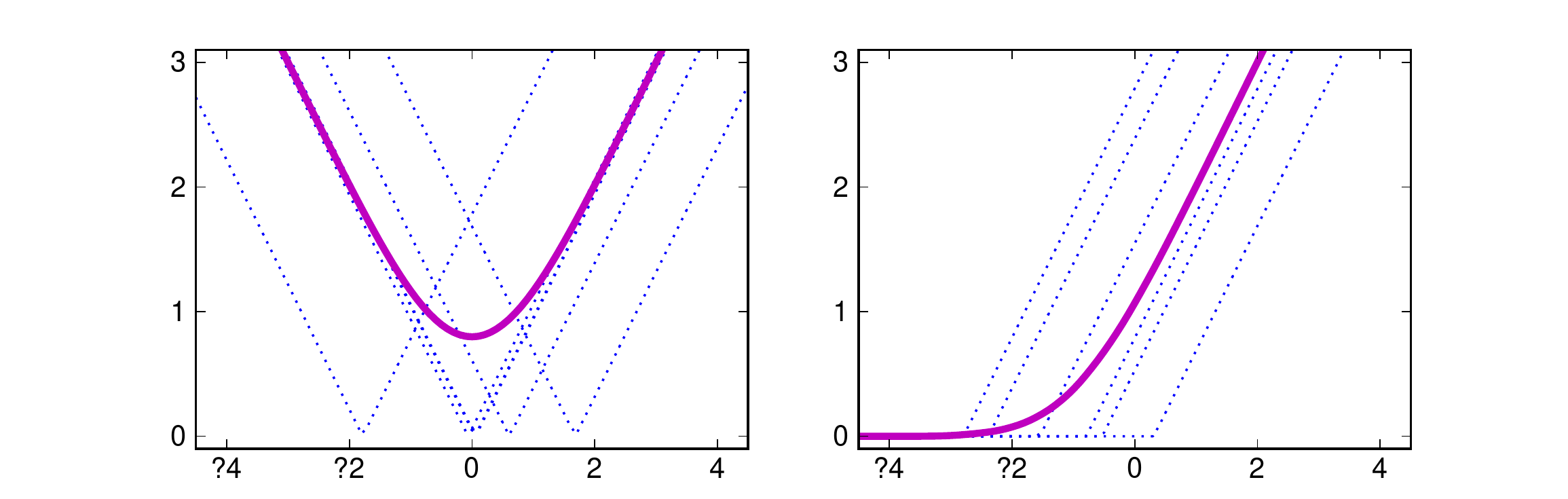}
\caption{Illustrating the expectation over non-smooth sample losses. 
In dotted blue, the loss functions for a few individual samples are shown,
each a non-smooth function. However, the expectation over a distribution
of such functions \emph{is} smooth, as shown by the thick magenta curve.
Left: absolute value, right: rectified linear function; samples are identical
but offset by a value drawn from $\Normal(0,1)$.
}
	\label{fig:absrect}
\end{figure*}

Many commonly used non-linearities (rectified linear units, absolute value normalization, etc.)
produce non-smooth sample loss functions.
However, when optimizing over a distribution of samples (or just a large enough dataset), 
the variability between samples can lead to a smooth expected loss function, even though
each sample has a non-smooth contribution. 
Figure~\ref{fig:absrect} illustrates this point for samples that have an absolute value or a rectified linear contribution to the loss.

It is clear from this observation that it is not possible to reliably estimate
the curvature of the true expected loss function, from the curvature of the individual
sample losses (which are all zero in the two examples above), if the sample losses are non-smooth.
This means that our previous approach of estimating the $h_i$ term in the 
optimal learning rate expression by a moving average of sample curvatures, as
estimated by the ``bbprop'' procedure \citep{lecun-98b} (which computes a Gauss-Newton approximation 
of the diagonal Hessian, at the cost of one additional backward pass) is 
limited to smooth sample loss functions, and we need a different approach for the 
general case\footnote{This also alleviates potential implementation effort, e.g., when using third-party software that does not implement bbprop.}.

\subsection{Finite-difference curvature}

A good estimate of the relevant curvature for our purposes (i.e., for determining a good learning rate)
is to not to compute the true Hessian at the current point,
but to take the expectation over noisy finite-difference steps,
where those steps are on the \emph{same scale} than the actually performed update steps, 
because this is the regime we care about.

In practice, we obtain this finite-difference estimates by computing two gradients of the same sample loss,
on points differing by the typical update distance\footnote{Of course, this estimate does not need to be computed at every step, which can save computation time.}:
\begin{equation}
h_i^{fd} = \left|\frac{\nabla_{\theta_i}- \nabla_{\theta_i+\delta_i}}{\delta_i}\right|
\label{eq:fd-sample}
\end{equation}
where $\delta_i = \hatg$.
This approach is related to the diagonal Hessian preconditioning in SGD-QN~\citep{bordes-jmlr-09}, but the step-difference used is different, and the moving average scheme there is decaying with time, 
which thus loses the suitability for non-stationary problems.



\begin{algorithm}[tb]
\DontPrintSemicolon
\SetKwInOut{Input}{input}
\SetKwInOut{Output}{output}
\label{alg:vSGD-fd}
\caption{vSGD-fd: minibatch-SGD with finite-difference-estimated adaptive learning rates }
 \Repeat{stopping criterion is met}{
  draw $\popsize$ samples,
  compute the gradients $\nabla_{\params}^{(j)}$ for each sample $j$\\ 
  compute the gradients on the same samples, with the parameters shifted by $\delta_i = \hatg$\\
  \For{$i \in \{1, \ldots,d\}$}{
  	compute finite-difference curvatures $h_i^{fd (j)} = \left|\frac{\nabla_{\theta_i}^{(j)}- \nabla_{\theta_i+\delta_i}^{(j)}}{\delta_i}\right|$ \\
  	\vspace{0.5em}
  	\If{$|\gradj - \hatg| > 2 \sqrt{\hatv-\hatg^2} \;\;\; \operatorname{or} \;\;\; \left|h_i^{fd} - \hath^{fd}\right| > 2 \sqrt{\hatv^{fd}-\left(\hath^{fd}\right)^2}$}
  	{\vspace{0.5em}
  	increase memory size for outliers $\tau_i \leftarrow \tau_i + 1$}{}
  	\vspace{0.5em}
  update moving averages
 $ \begin{array}{lll}
\hatg  &\leftarrow &(1-\tau_i^{-1}) \cdot \hatg + \tau_i^{-1} \cdot \frac{1}{\popsize}\sum_{j=1}^{\popsize}\gradj^{(j)}\\
\hatv  &\leftarrow &(1-\tau_i^{-1}) \cdot \hatv + \tau_i^{-1} \cdot \frac{1}{\popsize}\sum_{j=1}^{\popsize}\left(\gradj^{(j)}\right)^2\\
\hath^{fd}  &\leftarrow& (1-\tau_i^{-1}) \cdot \hath^{fd} + \tau_i^{-1} \cdot\frac{1}{\popsize}\sum_{j=1}^{\popsize} h_i^{fd (j)}\\
\hatv^{fd}  &\leftarrow& (1-\tau_i^{-1}) \cdot \hatv^{fd} + \tau_i^{-1} \cdot \frac{1}{\popsize}\sum_{j=1}^{\popsize}\left(h_i^{fd (j)}\right)^2\\
\end{array}$\\
\vspace{0.5em}
  estimate learning rate 
  $\;\;\displaystyle\eta_i^* \leftarrow \frac{\hath^{fd}}{\hatv^{fd}} \cdot \frac{\popsize \cdot(\hatg)^2}{\hatv + (\popsize-1)\cdot (\hatg)^2}$\\
  \vspace{0.5em}
 update memory size $\;\;\displaystyle\tau_i \leftarrow \left(1-\frac{(\hatg)^2}{\hatv}\right) \cdot  \tau_i+ 1 $\\
 \vspace{0.5em}
 update parameter $\;\; \theta_i \leftarrow \theta_i - \eta_i^*\cdot \frac{1}{\popsize}\sum_{j=1}^{\popsize}\gradj^{(j)}$\\
 }
 }
\end{algorithm}

\subsection{Curvature variability}
To further increase robustness, we reuse the same intuition that originally motivated 
vSGD, and take into account the variance of the curvature estimates 
(produced by the finite-difference method) to
reduce the likelihood of becoming overconfident (underestimating curvature, i.e., overestimating learning rates) by using a variance-normalization based on the signal-to-noise ratio of the curvature estimates.

For this purpose we maintain two additional moving averages:
\begin{eqnarray*}
\hath^{fd}  &\leftarrow& (1-\tau_i^{-1}) \cdot \hath^{fd} + \tau_i^{-1} \cdot h_i^{fd}\\
\hatv^{fd}  &\leftarrow& (1-\tau_i^{-1}) \cdot \hatv^{fd} + \tau_i^{-1} \cdot \left(h_i^{fd}\right)^2
\end{eqnarray*}
%
and then compute the curvature term simply as $\hath = \hatv^{fd}/\hath^{fd}$.

\subsection{Outlier detection}
If an  outlier sample is encountered
while the time constants $\tau_i$ is close to one (i.e., the history is mostly discarded from the moving averages at each update), this has the potential to disrupt the optimization process.
Here, the statistics we keep for the adaptive learning rates 
have an additional, unforeseen benefit: they make it trivial to detect outliers.

The outlier's effect can be mitigated relatively simply by increasing
the time-constant $\tau_i$ before incorporating the sample into the statistics
(to make sure old samples are not forgotten), and then due to
the perceived variance shooting up, the learning rate is automatically reduced.
If it was not an outlier, but a genuine change in the data distribution, the algorithm will quickly adapt, increase the learning rates again.

In practice, we use a detection threshold of two standard deviations, 
and increase the corresponding $\tau_i$ by one (see pseudocode).

\subsection{Algorithm}
Algorithm~\ref{alg:vSGD-fd} gives the explicit pseudocode for this finite-difference
estimation, in combination with the minibatch size-adjusted rates from equation~\ref{eq:opt-lrmb},
termed ``vSGD-fd''.
Initialization is akin to the one of vSGD, in that all moving averages are bootstrapped on a few samples (10) before any updates are done.
It is also wise to add an tiny $\epsilon=10^{-5}$ term where necessary to avoid divisions by zero.

\begin{figure*}
	\centering
		\includegraphics[width=0.8\textwidth]{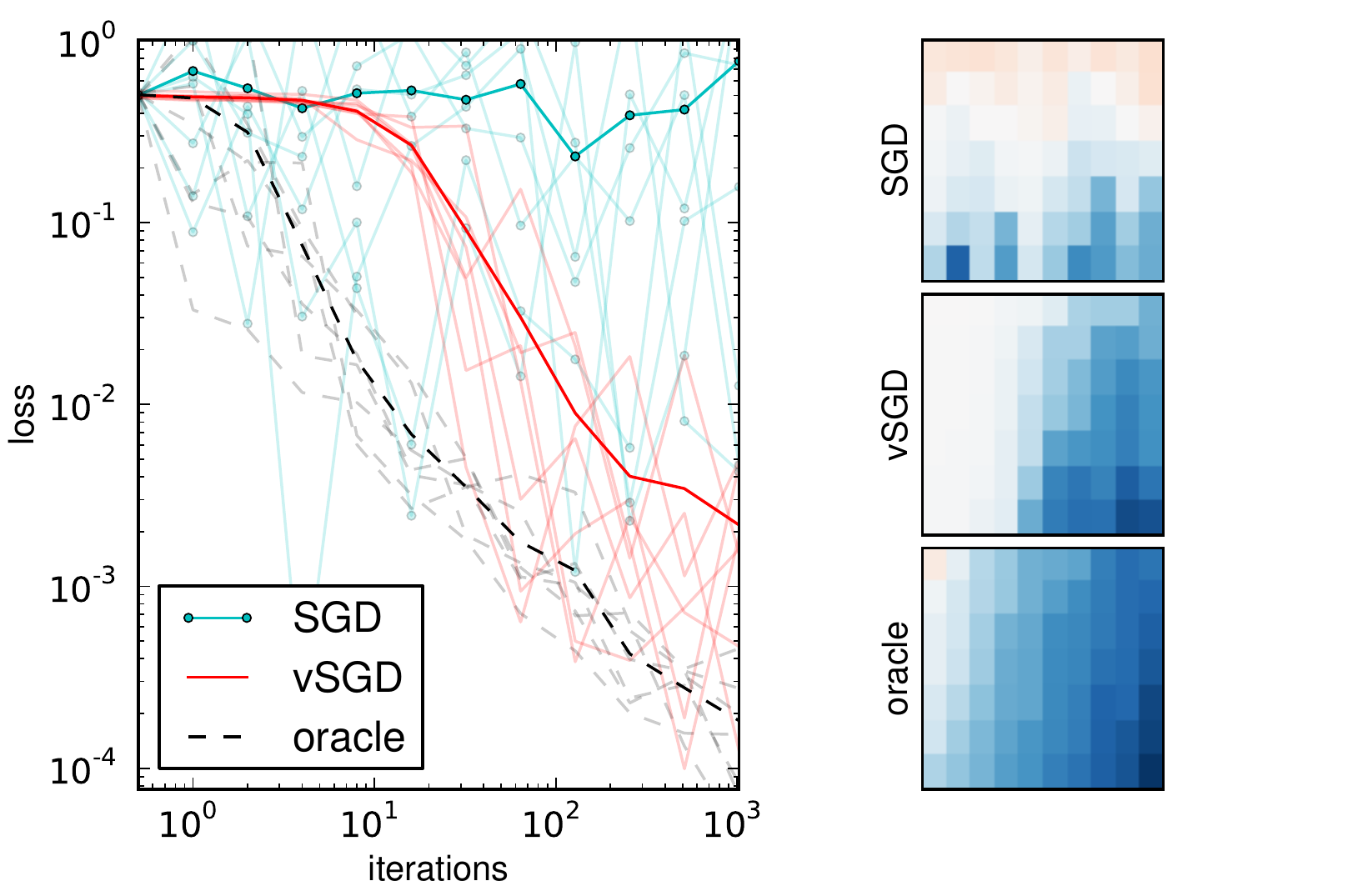}
\caption{Explanation of how to read our concise heatmap
performance plots (right), based on the more common representation 
as learning curves (left).
In the learning curve representation, we plot one curve for 
each algorithm and each trial (3x8 total), with a unique color/line-type
per algorithm, and the mean performance per algorithm with more contrast.
Performance is measured every power of 2 iterations.
This gives a good idea of the progress, but becomes quickly hard to read.
On the right side, we plot the identical data in heatmap format. Each 
square corresponds to one algorithm, the horizontal axis are still the 
iterations (on $\log_2$ scale), and on the vertical axis we arrange (sort)
the performance of the different trials at the given iteration.
The color scale is as follows: white is the initial loss value,
the stronger the blue, the lower the loss, and if the color is reddish, the algorithm
overjumped to loss values that are bigger than the initial one.
Good algorithm performance is visible when the square becomes blue on the right side,
instability is marked in red, and the variability of the algorithm across trials is visible 
by the color range on the vertical axis.
}
	\label{fig:pmapping}
\end{figure*}

\section{Simulations}

An algorithm that has the ambition to work out-of-the-box, without any tuning of 
hyper-parameters, must be able to pass a number of elementary tests: those may not be
sufficient, but they are necessary.
To that purpose, we set up a collection of elementary (one-dimensional) stochastic
optimization test cases, varying the shape of the loss function,
its curvature, and the noise level.
The sample loss functions are
\begin{eqnarray*}
f_{quad} &=& A\cdot \left(\theta-\xi^{(j)}\right)^2\\
f_{abs} &=& A\cdot\left|\theta-\xi^{(j)}\right|\\
f_{rectlin} &=& \left\{\begin{array}{ll}
	A \cdot (\theta-\xi^{(j)}) & \operatorname{if} \theta-\xi^{(j)}>0\\
	0 & \operatorname{otherwise}
\end{array}\right.\\
f_{gauss} &=& A - A e^{-\frac{\left(\theta-\xi^{(j)}\right)^2}{2}}\\
\label{eq:funs}
\end{eqnarray*}
where $A$ is the curvature setting and the $\xi^{(j)}$ are drawn from $\Normal(0, \sigma^2)$.
We vary curvature and noise levels by two orders of magnitude, i.e.,
$A \in \{0.1, 1, 10\}$ and $\sigma^2 \in \{0.1, 1, 10\}$, giving us 9x4 test cases.
To visualize the large number of results,
we summarize the each test case and algorithm combination in a concise
heatmap square (see Figure~\ref{fig:pmapping} for the full explanation).

\begin{figure*}
	\centerline{
		\includegraphics[width=1.01\textwidth]{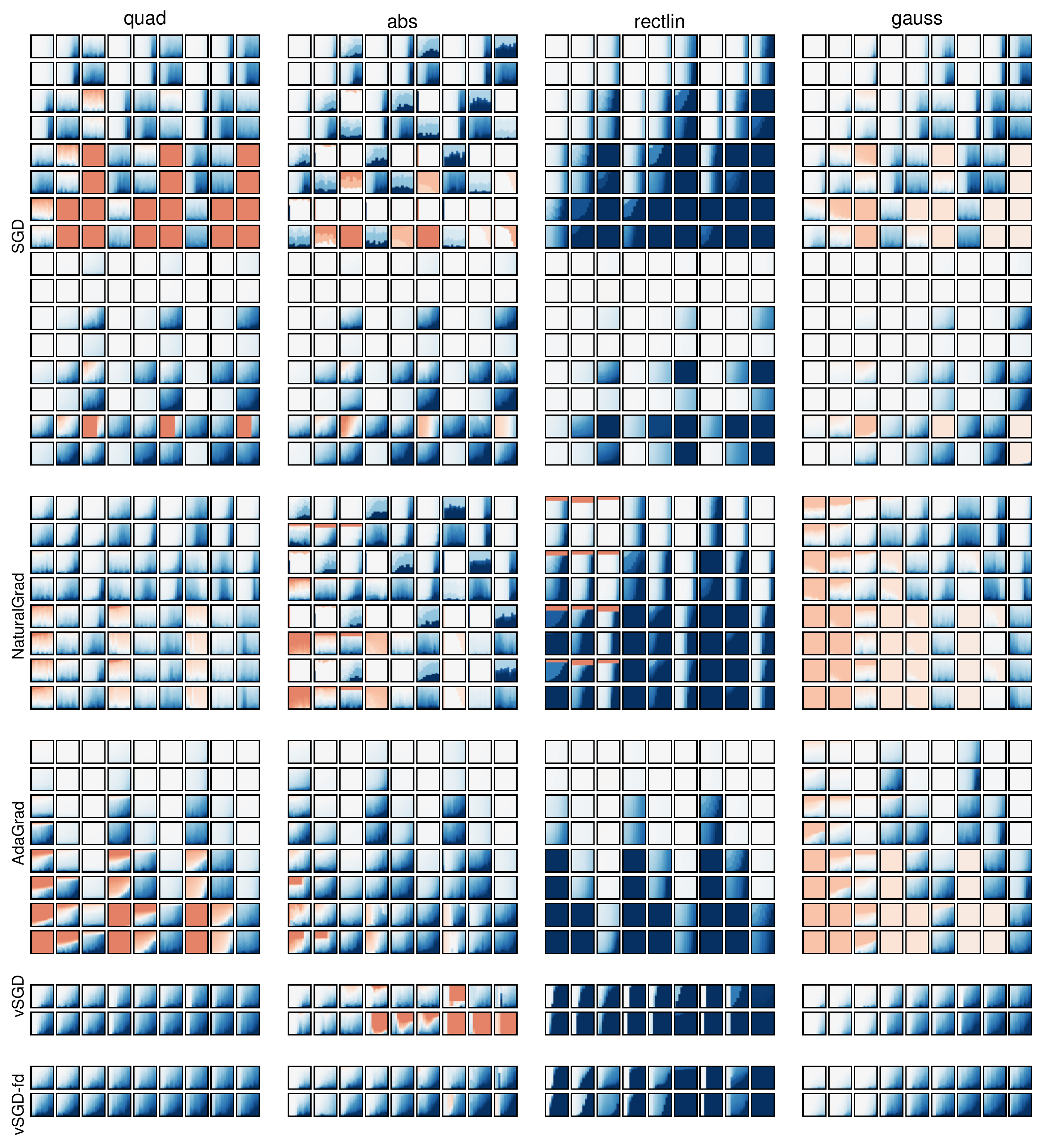}
}
\caption{
Performance comparisons for a number of algorithms (row groups) under different setting variants (rows) and sample loss functions (columns), the latter grouped by loss function shape.
Red tones indicate a loss value worsening from its initial value, white corresponds to no progress, 
and darker blue tones indicate a reduction of loss (in log-scale).
For a detailed explanation of how to read the heatmaps, see Figure~\ref{fig:pmapping}.
The new proposed algorithm vSGD-fd (bottom row group) 
performs well across all functions and noise-level settings, namely fixing the vSGD instability on
non-smooth functions like the absolute value.
The other algorithms need to have their hyper-parameters tuned to the task to work well.
}
	\label{fig:variants}
\end{figure*}

In Figure~\ref{fig:variants}, we show the results for all test cases on a range of algorithms and minibatch sizes $n$.
Each square shows the gain in loss for 100 independent runs of 1024 updates each.
Each group of columns corresponds to one of the four functions, with the 9 inner columns using different curvature and noise level settings. Color scales are identical for all heatmaps within a column, but not across columns.
Each group of rows corresponds to one algorithm, with each row using a different hyper-parameter setting,
namely initial learning rates $\eta_0 \in \{0.01, 0.1, 1, 10\}$ (for SGD, \adagrad~\cite{DuchiHS11} and the natural gradient~\cite{amari2000adaptive}) and decay rate $\gamma \in \{0,1\}$ for SGD.
All rows come in pairs, with the upper one using pure SGD ($\popsize=1$) and the lower one using minibatches ($\popsize=10$).

The findings are clear: in contrast to the other algorithms tested, vSGD-fd does not require any hyper-parameter tuning to 
give reliably good performance on the broad range of tests: the learning rates adapt automatically to different curvatures and
noise levels. And in contrast to the predecessor vSGD, it also deals with non-smooth loss functions appropriately. 
The learning rates are adjusted automatically according to the minibatch size, which improves convergence speed on the noisier 
test cases (3 left columns), where there is a larger potential gain from minibatches.

The earlier variant (vSGD) was shown to work very robustly on a broad range of real-world benchmarks and
non-convex, deep neural network-based loss functions. We expect those results on smooth losses to transfer directly 
to vSGD-fd. This bodes well for future work that will determine its performance on real-world non-smooth problems.

\section{Conclusion}

We have presented a novel variant of SGD with adaptive learning rates
that expands on previous work in three directions. 
The adaptive rates properly take into account the minibatch size,
which in combination with sparse gradients drastically alleviates the
diminishing returns of parallelization. Also, the curvature estimation procedure is based on a finite-difference approach that 
can deal with non-smooth sample loss functions.
The final algorithm integrates these components, has linear complexity and is hyper-parameter free.
Unlike other adaptive schemes, it works on a broad range of elementary test cases,
the necessary condition for an out-of-the-box method.

Future work will investigate how to adjust the presented element-wise approach
to highly nonseparable problems (tightly correlated gradient dimensions), 
potentially relying on a low-rank or block-decomposed estimate of the gradient covariance matrix,
as in TONGA~\cite{leroux-nips-08}.

\subsubsection*{Acknowledgments}

The authors want to thank Sixin Zhang, Durk Kingma, Daan Wierstra, Camille Couprie, Cl\'{e}ment Farabet and
Arthur Szlam for helpful discussions. We also thank the reviewers for helpful suggestions, and the `Open Reviewing Network' for perfectly managing the novel open and transparent reviewing process.
This work was funded in part through AFR
postdoc grant number 2915104, of the National Research Fund
Luxembourg.



\bibliographystyle{unsrtModif}
\bibliography{sgd,more}

\end{document}